\documentclass[runningheads]{llncs}
\usepackage[T1]{fontenc}
\usepackage{graphicx}
\usepackage{booktabs}
\usepackage{makecell}
\usepackage{xcolor}
\usepackage{hyperref}
\usepackage{paralist}
\usepackage[misc]{ifsym}
\newcommand{\corr}{(\Letter)}

\usepackage{xspace}
\newcommand{\ourdata}{NRITYAM\xspace}

\begin{document}

\title{NRITYAM: Language Models Meet Art and Heritage of Dance}

\titlerunning{{NRITYAM}: Language Models Meet Art and Heritage of Dance}

\author{Punit Kumar Singh\inst{1,5} \and
Niladri Ghosh\inst{4} \and
Advait Joshi\inst{6} \and Shailee Choudhary\inst{2}\and Michael F\"arber\inst{3}\and Haiqin Yang \corr \inst{1, 7} }

\authorrunning{P. K. Singh, N. Ghosh, A. Joshi, S. Choudhary, M. F\"arber, and H. Yang}

\institute{Shenzhen Technology University, 518118 Shenzhen, China 
\email{{yanghaiqin}@sztu.edu.cn}
\and
New Delhi Institute of Management, New Delhi, India 
\and
Technische Universität Dresden, 01069 Dresden, Germany
\and
Ramakrishna Mission Vivekananda Educational and Research Institute, India
\and
Indian Institute of Technology, India
\and
Swami Vivekananda Institute of Technology, India
\and GuangDong Engineering Technology Research Center of Edge Intelligence
}

\maketitle              

\begin{abstract}
Language models have become essential tools in shaping modern workflows. However, their global effectiveness hinges on a nuanced understanding of local socio-cultural contexts. To address this gap, we present NRITYAM, a comprehensive benchmark for evaluating the cultural comprehension capabilities of language models in the context of global dance traditions. NRITYAM comprises 9,260 carefully curated question-answer pairs spanning 12 languages, making it the largest dataset dedicated to evaluating cultural knowledge in dance. The dataset has been developed from the ground up through close collaboration with native dance artists and native speakers of the languages, who authored and validated culturally relevant questions specific to their regions. We evaluate a broad set of models, including large language models, small language models, multimodal large language models, and small multimodal language models. As a multilingual and multicultural benchmark, NRITYAM sets a new standard for evaluating the ability of AI systems to understand and reason about traditional performing arts. Detailed dataset samples are available at~\url{https://github.com/niladrighosh03/NRITYAM}.

\end{abstract}
\setcounter{footnote}{0}
\section{Introduction}
Dance serves as a powerful medium for cultural and emotional expression, bringing together individuals from varied backgrounds and traditions \cite{bannerman2014dance}.  Exploring different traditional dance forms offers meaningful perspectives on the values, historical experiences, and social frameworks of the communities that create and uphold them \cite{peng2024historical}. Moreover, dance holds significant societal importance, serving as a medium for preserving cultural knowledge and shaping collective identity \cite{afolaranmi2024cultural}. The language, rituals, and evolving forms of dance serve as rich expressions of a community’s historical experiences, cultural transformations, and collective identity encompassing elements such as traditional attire, religious associations, embodied movement, and social heritage~\cite{desmond1993embodying}.\par

\begin{figure}[t]
\centering 
\includegraphics[width=.9\textwidth]{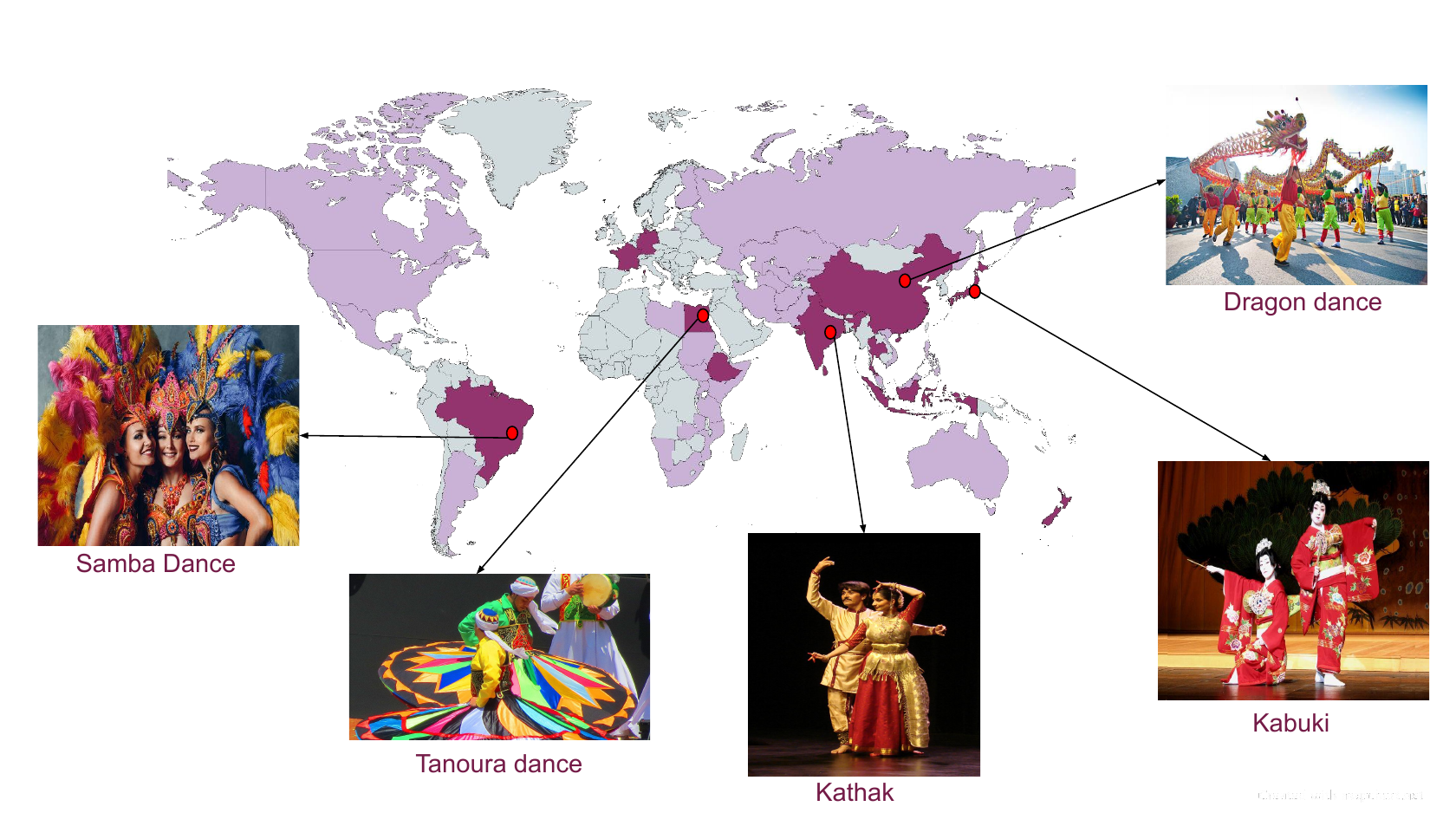}
\caption{ \ourdata is a diverse benchmark featuring 12 languages, with questions manually created and verified by native language speakers and dance experts.  It spans 8 key aspects of traditional dance across two modalities, text and image, emphasizing mid- to low-resource languages.  The benchmark features dances originating from 12 countries across 5 continents that are now performed in 100 countries across 6 continents, which are visualized with dark blue for their origins and light blue for their current reach. \ourdata offers a wide range of question formats, including multiple-choice questions (MCQs) and both short and long visual question-answering (VQA) tasks.
} 
\label{World}
\end{figure}

Researchers have increasingly turned to traditional dance as a rich lens for analyzing cultural dynamics, providing a robust framework for understanding how embodied practices evolve across regional and societal boundaries \cite{lucchi2025dance}. Although many traditional dance forms share foundational elements such as expressive movement, symbolic gestures, and rhythmic structure, they manifest uniquely within different cultural contexts. These divergences are evident in variations in movement style, facial expression, ritual significance, performance settings, and the terminology used to describe them. For instance, Bharatanatyam, a classical Indian dance rooted in Tamil Nadu, has been adapted in Sri Lanka through a distinctive fusion with Kandyan dance aesthetics, reflecting localized narratives and performance traditions. Such adaptations highlight how traditional dance not only preserves the heritage but also evolves through intercultural exchange and regional reinterpretation. Language Models (LMs) have revolutionized natural language understanding, content generation, and decision-making, becoming indispensable across industries such as education, governance, and entertainment \cite{brown2020language}. From Large Language Models (LLMs) to Multimodal Language Models (MLMs) and Small Language Models (SLMs)~\footnote{Any model with 7B parameters or fewer is considered a Small Language Model (SLM) in this work.}, these advancements have enabled seamless communication and efficient problem-solving \cite{ouyang2022training}. However, a persistent challenge remains: ensuring that these models effectively recognize and reason about diverse linguistic and cultural contexts, particularly in underrepresented domains such as traditional dance \cite{bender2021dangers}.\par

Traditional and indigenous dance forms are deeply embedded in local histories, societal values, and cultural identities~\cite{guo2025impact}. Despite their significance, current language models are predominantly trained and evaluated on global popular culture and mainstream dance styles, such as hip-hop, often neglecting heritage dance traditions and culturally distinctive practices.  This bias can perpetuate inaccuracies, stereotypes, and the marginalization of underrepresented communities.  Conversely,  models capable of understanding and respecting cultural nuances can enhance performance while promoting greater inclusivity and equity in AI applications.

\textbf{Motivation for \ourdata\footnote{The Sanskrit term ``NRITYAM'' holds profound significance in classical Indian dance.} Dataset.} Existing dance-related benchmarks are largely monolingual, English-centric~\cite{aristidou2022safeguarding,ma2024vimva}, and focused primarily on motion recognition~\cite{grammalidis2016treasures}.  For example, PopDanceSet~\cite{luo2024popdg} is a monolingual dataset.  To date, no comprehensive benchmark captures the rich cultural nuances of traditional dance reasoning across multiple languages, diverse cultural contexts, and visual question answering (VQA).  To address this gap, we introduce \ourdata, the \textbf{largest multicultural and multilingual traditional dance benchmark} to date.  It consists of approximately 9,260 dance-related questions covering traditional dances originating from 12 countries across 5 continents, which are now performed in over 100 countries spanning 6 continents.  The benchmark evaluates the capabilities of LLMs, SLMs, and MLMs across twelve languages.  Questions are organized into two modalities, text-based and image-based, and systematically categorized into three key types: history-based, rule-based, scenario-based.~\footnote {MLMs are used exclusively for evaluating the image-based questions.} 

This work is guided by the following research questions: 
\begin{compactenum}[(1)]
  \item How do different categories of models, i.e., LLMs, SLMs, MLMs, and Small Multimodal Language Models (SMLMs), perform on the \ourdata dataset? 
  \item What trends and patterns emerge in model performance across the various question types, including history-based, rule-based, scenario-based, and image-based questions, in the \ourdata dataset?

  \item What are the performance trends of language models across different countries or languages in Asia, Africa, South America, Oceania, and Europe?
\end{compactenum}

\textbf{Key contributions.}  Our main contributions are as follows:

\textbf{1. \textbf{The {NRITYAM}} Dataset:}  We present the first and the most comprehensive QA dataset on traditional dance, covering 12 countries across 5 continents (with global reach across 100+ countries on 6 continents).  The dataset is available in 12 native languages as well as in English.


\textbf{2. Diverse Question Types:} The dataset includes 9,260 questions spanning two modalities (text and image) and three categories, challenging AI models to reason over textual, visual, multilingual, and culturally grounded inputs.

\textbf{3. Comprehensive Benchmarking:} We rigorously evaluate 13 state-of-the-art language models, including LLMs, SLMs, MLMs, and SMLMs, uncovering critical gaps in their cultural and contextual reasoning abilities regarding traditional dance.

By addressing cultural under-representation in AI, \textbf{NRITYAM} establishes a robust benchmark for evaluating and improving AI systems. This research advances the intersection of NLP and culturally rich domains, contributing to greater inclusivity and equity in AI applications worldwide.

\section{Related Work}
\textbf{Prior Cultural VQA Benchmarks.} Pioneering efforts in culturally aware Vision-Question Answering (VQA) have led to datasets spanning various regional and thematic dimensions, including FM-IQA~\cite{gao2015you}, MCVQA~\cite{gupta2020unified}, xGQA~\cite{pfeiffer2022xgqa}, MaXM~\cite{changpinyo2023maxm}, MTVQA~\cite{tang2025mtvqa}, MABL~\cite{kabra2023multi}, MAPS~\cite{liu2024multilingual}, and MaRVL~\cite{liu2021visually}.  Parallel lines of work, such as CVQA~\cite{mogrovejo2024cvqa}, CulturalVQA~\cite{nayak2024benchmarking}, ALM-bench~\cite{vayani2024all}, and CultSportQA~\cite{singh2025let}, offer broader resources covering distinct cultural themes, with CVQA introducing multilingual queries paired with English translations.  Other benchmarks adopt granular geographic or thematic scopes; for instance, SEA-VQA~\cite{urailertprasert2024sea} focuses strictly on Southeast Asia, while FoodieQA~\cite{li2024foodieqa}, World Wide Dishes~\cite{magomere2024you}, and WORLDCUISINES~\cite{winata2025worldcuisines} focus on specific culinary traditions.  While \ourdata shares the foundational objective of using a specific cultural lens, i.e., traditional dance, it distinctly expands upon prior art through a substantially larger dataset and broader multilingual coverage.

\textbf{Sociocultural Reasoning in LLMs.} Complementing visual benchmarks, recent NLP research scrutinizes the behavioral and alignment gaps of LLMs through established sociological frameworks like the World Values Survey and Hofstede’s cultural dimensions~\cite{DBLP:conf/acl/AlKhamissiEAD24}.  These studies frequently reveal critical alignment deficiencies when adapting systems to user-specific or non-Western contexts~\cite{johnson2022ghost}. Although interventions like synthetic personas and targeted fine-tuning show promise in elevating cultural adaptability and safe text moderation (e.g., cross-lingual hate speech detection), regional and low-resource language performance persistently lags behind English baselines~\cite{kwokevaluating,dwivedi2024exploring,deng2024multilingual}. These systemic limitations underscore the critical necessity for robust, multilingual evaluation suites to genuinely foster and measure the cultural competence of large foundation models.

\begin{table*}[t]
\caption{Comparison of our dataset with other dance-related datasets.
The metadata compared includes the number of samples (questions), the number of dances, whether cultural aspects are considered, the number of languages, modalities (i.e., whether the data includes multimodal questions), and question type.
``\textbf{---}'' indicates the value is not specified.}
\label{tab:dataset-comparison}
\centering
\scalebox{0.74}{
\setlength{\tabcolsep}{6pt}
\renewcommand{\arraystretch}{1.3}
\begin{tabular}{@{} l c c c c l l @{}}
\toprule
\textbf{Dataset} &
\textbf{\#Samples} &
\textbf{\#Dances} &
\textbf{\makecell{Cultural\\Aspects}} &
\textbf{\#Languages} &
\textbf{Modalities} &
\textbf{\makecell[l]{Question\\Type}} \\
\midrule
CULTURALVQA~\cite{nayak2024benchmarking} &
2{,}378 & --- & No & 20 & Image & MCQ \\

CVQA~\cite{mogrovejo2024cvqa} &
10{,}000 & --- & No & 31 & Text + Image & MCQ \\

\textbf{\textit{NRITYAM (ours)}} &
\textbf{9{,}260} & \textbf{30} & \textbf{Yes} & \textbf{12} &
Text + Image & \textbf{MCQ} \\
\bottomrule
\end{tabular}
}
\end{table*}

\section{Construction of {NRITYAM }}

\subsection{Manual Data Collection}
The creation of \ourdata follows a carefully structured, multi-phase process to ensure comprehensive coverage and high-quality standards.  Domain experts and country-specific annotators contribute at every stage, from data collection to question formulation and manual translation across multiple languages, incorporating their cultural knowledge and expertise

\textbf{Data Sources:} The dataset is built by sourcing information from multiple diverse and reliable platforms, including Wikipedia, government culture websites, local culture dance blogs, dance culture journals, and news outlets. These sources are carefully selected to authentically capture the essence of traditional dance from 12 countries: Brazil, China, Egypt, Ethiopia, France, Germany, India, Indonesia, Japan, New Zealand, Sri Lanka, Thailand, with emphasis on their cultural and regional significance.  The questions are designed to focus on historical, rule-based, and image-based aspects of the dance.

\textbf{1. Wikipedia:} As a foundational resource, Wikipedia offers well-documented, detailed insights into the history, origins, and rules of traditional dance, serving as a cornerstone for verified information across diverse dance traditions.

\textbf{2. Government Culture Website:} The government culture websites of each country contribute unique perspectives on the historical and cultural relevance of traditional dance forms, ensuring authenticity and depth.

\textbf{3. Local Culture Dance Blogs:} Dedicated blogs and websites focusing on indigenous and traditional dance offer community-driven perspectives, unique insights, and region-specific practices associated with each dance form.

\textbf{4. Dance Cultural Journals:} Scholarly journals and publications specializing in cultural studies enrich the dataset with detailed articles on the evolution and societal impact of traditional dance across different regions.

\textbf{5. News outlets:} Renowned news outlets from each country provide contemporary coverage, including recent trends, key events, and ongoing efforts to preserve or revive traditional dances.

\textbf{Dataset Organization}

The NRITYAM dataset is divided into two categories: text-based and image-based.  Each category is further organized into three question types: history-based, rule-based, and scenario-based questions.

The dataset follows a multiple-choice question (MCQ) format with four options (A, B, C, D), out of which only one is correct.  Each question-answer (QA) pair includes metadata such as continent, country, posture, religion, and other question types.
\begin{figure*}
\centering 
\includegraphics[width=0.7\textwidth]{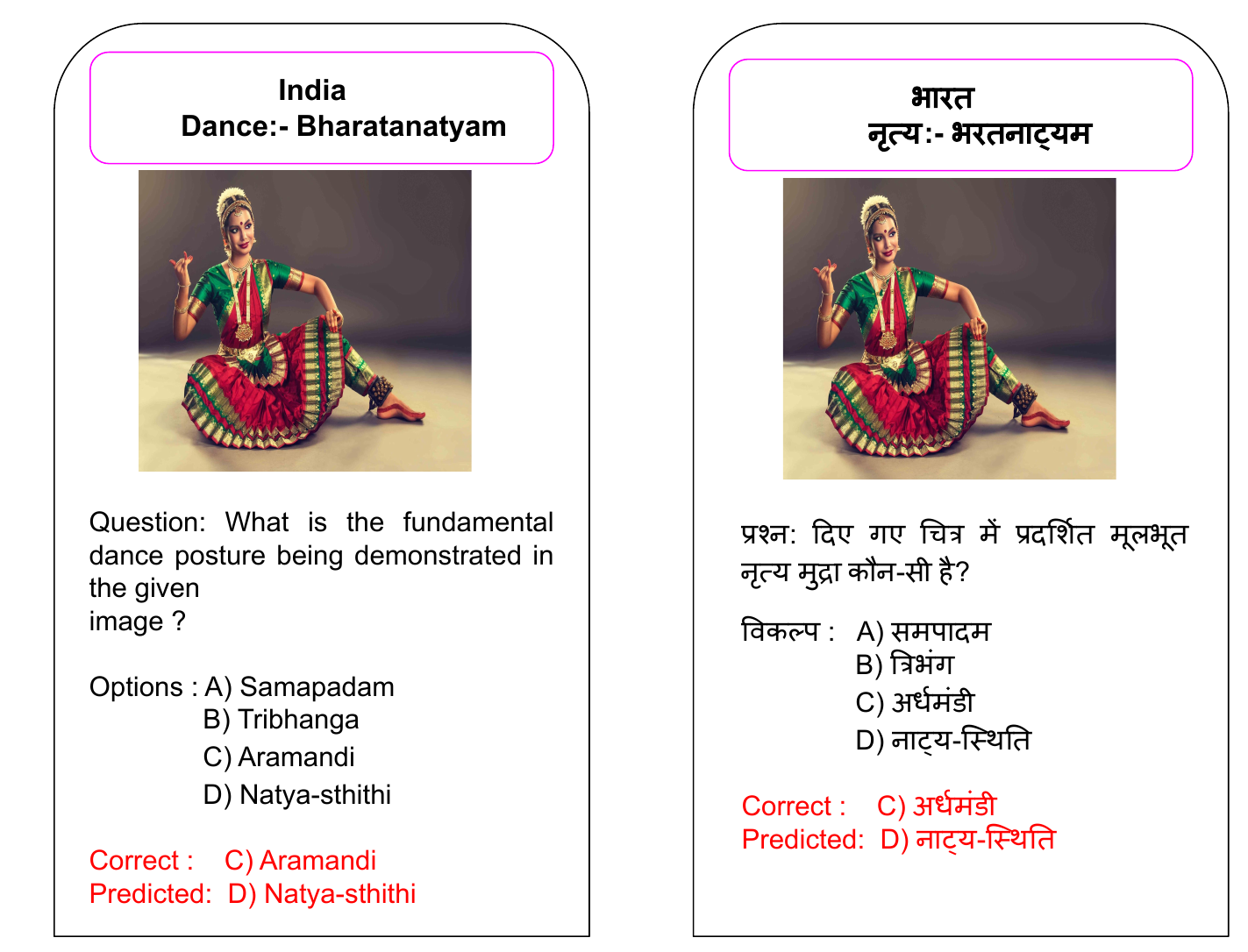}
\caption{Example illustration of india traditional dance wrong prediction by language model (in hindi and english)} 
\label{Hindi1}
\end{figure*}

\begin{figure*}
\centering 
\includegraphics[width=0.7\textwidth]{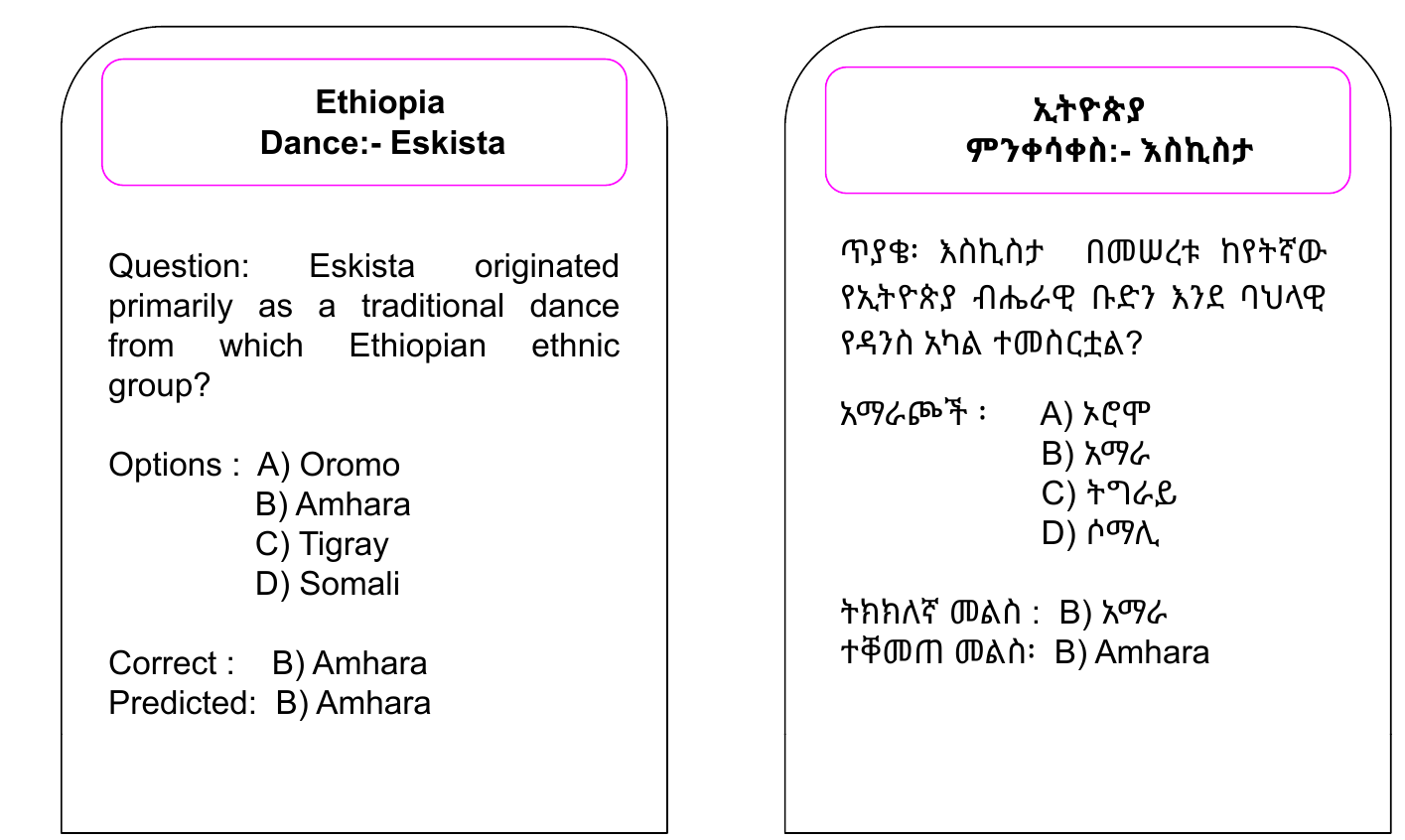}
\caption{Example illustration of ethiopia traditional dance correct prediction by language model (in amharic and english)} 
\label{Ethiopia}
\end{figure*}
The text-based questions are evaluated using LLMs and SLMs, while the image-based questions are assessed using MLMs and SMLMs.

History-based questions test the model’s knowledge of a dance's origins and cultural significance. Scenario-based questions assess the model’s ability to determine the most appropriate move in a given dance situation for a good performance. Rule-based questions evaluate the model’s understanding of the fundamental rules of the dance depicted in the text or image. Religion-based questions assess the model’s understanding of dances associated with particular religions, and so on.  Examples from the NRITYAM dataset are illustrated in Figures \ref{Hindi1} and \ref{Ethiopia}.

\subsection{Annotation Process}

We outline the main steps in the annotation below:

\textbf{1. Team Structure and Annotators Background.} We hire 36 expert workers from 12 countries, with 3 representatives from each country, based on the following eligibility criteria: (1) The individual must be a native speaker. (2) They must have lived in the country for at least 10 years.
(3) They must possess a strong understanding of local culture and traditional dance.
(4) Their parents must also be from the country and currently reside there. (5) They must hold at least a degree from a nationally accredited dance institution. Two-thirds of the annotators are responsible for creating questions based on the provided guidelines, leveraging their knowledge of traditional dance.  The remaining annotator was tasked with validating and filtering out questions that fail to meet quality standards.\par

\begin{figure*}
\centering 
\includegraphics[width=0.9\textwidth]{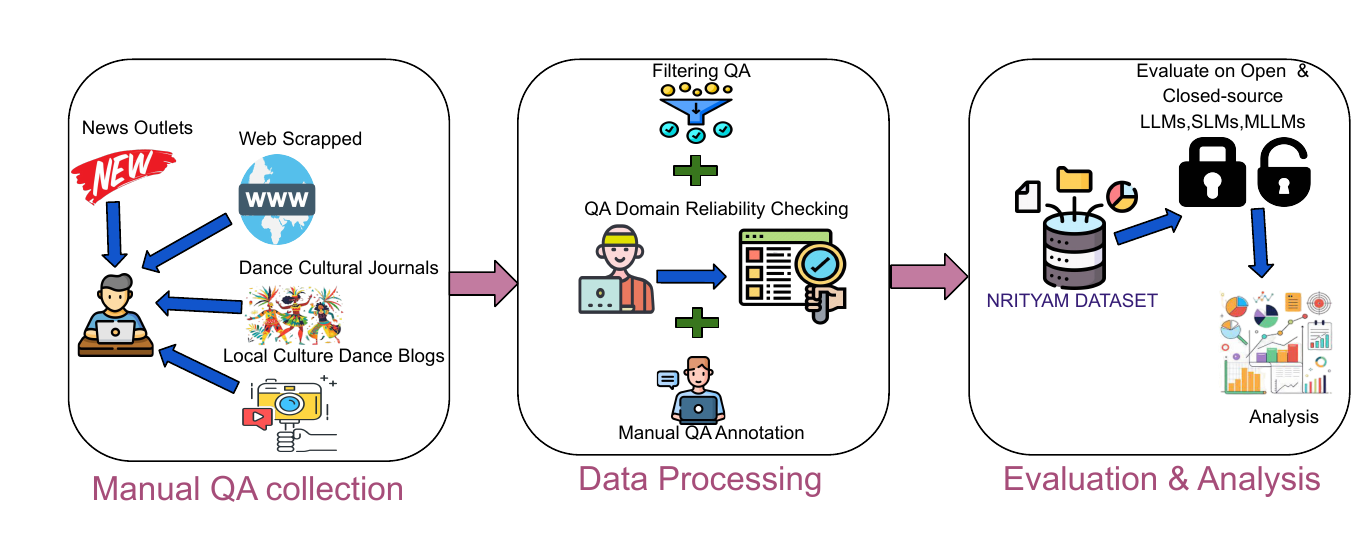}
\caption{Manual dataset construction pipeline for \ourdata: The data collection process involved two key stages: (1) annotators gather data sources and generate questions, drawing from their respective cultural backgrounds and languages; (2) annotators review and verify the questions to ensure cultural authenticity and maintain high translation quality.} 
\label{Intron1}
\end{figure*}

\textbf{2. Question Formation.}
For each selected textual passage or image, the annotator’s first task is to verify whether the content aligned with the traditional dance associated with the country. Content unrelated to the traditional dance is immediately rejected. If the content is relevant, the annotator creates questions focusing on rules, history, location, attire, religion, scenario, artist, and posture.  Each question has to be complete, self-contained, and understandable without additional context.  The questions follow a multiple-choice format, consisting of four options, with only one correct answer.  The final annotated format includes the source passage, a relationship attribute indicating the question’s context, the type of question (e.g., history-based, rule-based, scenario-based), and the four answer options.  After a question is constructed, it is translated into the regional language that the annotator is familiar with.  The annotators are paid at a rate between \$0.10 and \$0.50 per example, depending on the country exchange rate and difficulty of annotation.

\textbf{3. Training and Guidelines.} Annotators are receiving comprehensive training that covers the objectives of the \ourdata dataset, clear definitions and examples of various question types, and best practices for ensuring consistency and cultural sensitivity.  Detailed guideline documents are being provided, including templates, metadata tagging standards, and examples of culturally appropriate representations.  Additional sessions are focusing on language and cultural training, emphasizing the correct use of local terminologies and traditions.  Each participant is required to attend a live one-hour online workshop or watch a recorded version. These sessions are offering an overview of the project, explaining the task instructions in detail, and addressing any potential questions. To ensure a thorough understanding of the assignment, a pilot study is being conducted before the main annotation phase.

\textbf{4. Quality Assurance and Cross-Validation.} A rigorous quality assurance process is being conducted through multi-step validations.  Each question-answer pair is undergoing cross-validation by at least one annotator, who understands the basic requirements to be qualified for inclusion in the dataset, as well as the translation quality.  Image-based questions are being reviewed for proper alignment between visual elements and textual prompts.  Spot checks and random sampling are being performed by quality analysts to maintain clarity and consistency.  Bias mitigation measures are ensuring a balanced representation of dance across regions and question types, while cultural sensitivity reviews are eliminating stereotypes or offensive content.

\section{Statistical Analysis of {NRITYAM}}

\begin{figure*}
\centering
\includegraphics[width=.9\textwidth]{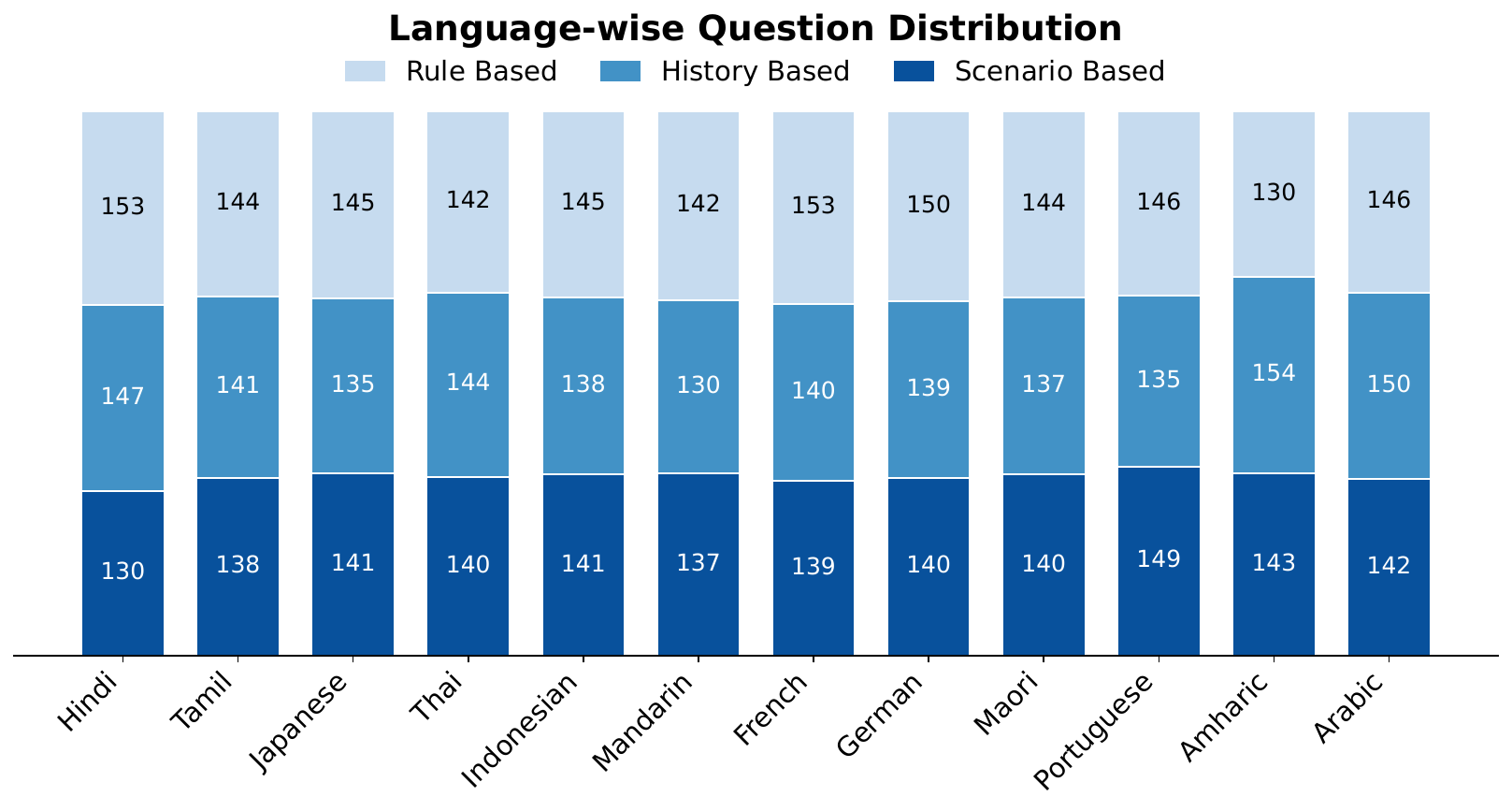}
\caption{\textbf{Distribution of text-based questions across language.}} 
\label{Text1}
\end{figure*}

\begin{figure*}
\centering
\includegraphics[width=.9\textwidth]{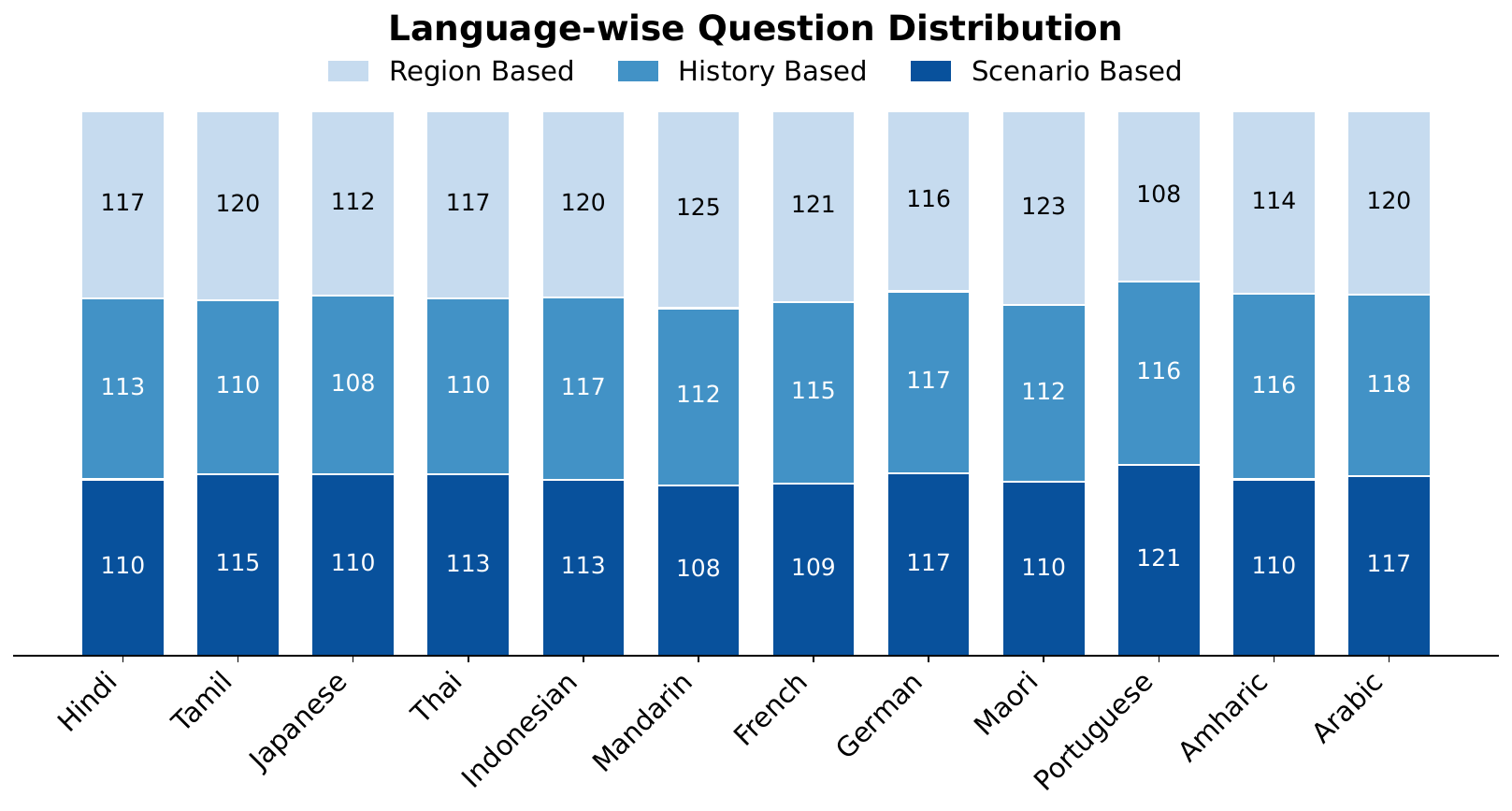}
\caption{\textbf{Distribution of image-based questions across language.}} 
\label{image}
\end{figure*}

The \ourdata dataset shows a balanced mix of text-based and image-based questions, with a slight dominance of text-based ones comprising 5,110 questions over the visual ones, which comprise of 4,150. Image-based questions mainly focuses on dance rules, scenarios, and history. Figures~\ref{Text1} and ~\ref{image} show the distribution of text and image-based questions across languages. 

\section{Experiments}
\subsection{Models}
To comprehensively evaluate our proposed benchmark, \ourdata, we conduct an extensive assessment across a diverse suite of foundation models spanning both textual and visual modalities.  For text-based evaluation, our suite encompasses leading LLMs, including Llama-3.1~\cite{dubey2024llama}, GPT-4~\cite{achiam2023gpt}, GPT-5~\cite{singh2025openai}, Gemma-7B~\cite{team2024gemma}, Phi-4~\cite{abdin2024phi}, Qwen-3~\cite{bai2023qwen}, and Claude-4.5.
Beyond text-based models, we evaluate a wide range of MLMs to assess dance reasoning capabilities in multilingual settings.  Specifically, this visual category includes mBLIP (a BLIP-2-based model)~\cite{geigle2023mblip}, PaliGemma-2~\cite{beyer2024paligemma}, LLaVA~\cite{liu2023visual}, Qwen3-VL~\cite{bai2025qwen3}, and DeepSeek-OCR~\cite{wei2025deepseek}.
\subsection{Evaluation Setup}
We conducted a comprehensive evaluation of the \ourdata dataset, which comprises text and image-based {Multiple Choice Questions (MCQs)} categorized into three distinct dimensions: {1. Cultural and Historical Knowledge}, {2. Rule Comprehension}, and {3. Scenario-Based Reasoning}.  To systematically benchmark model performance across diverse languages and modalities, all evaluations are executed under a \textbf{zero-shot} prompting paradigm.  For reproducibility and consistency, the {temperature is set to 0}, and \textbf{accuracy} serves as the primary evaluation metric.  For deployment, {open-source models} are initialized using {16-bit floating-point precision (FP16)} and evaluated via {greedy decoding}, whereas {proprietary models} are queried through their official developer APIs. Final model predictions are extracted based on the highest output token probability, establishing a standardized and deterministic evaluation pipeline.


\section{Discussion on Results}

\subsection{Main Results} 
\begin{figure*}
\includegraphics[width=1\textwidth]{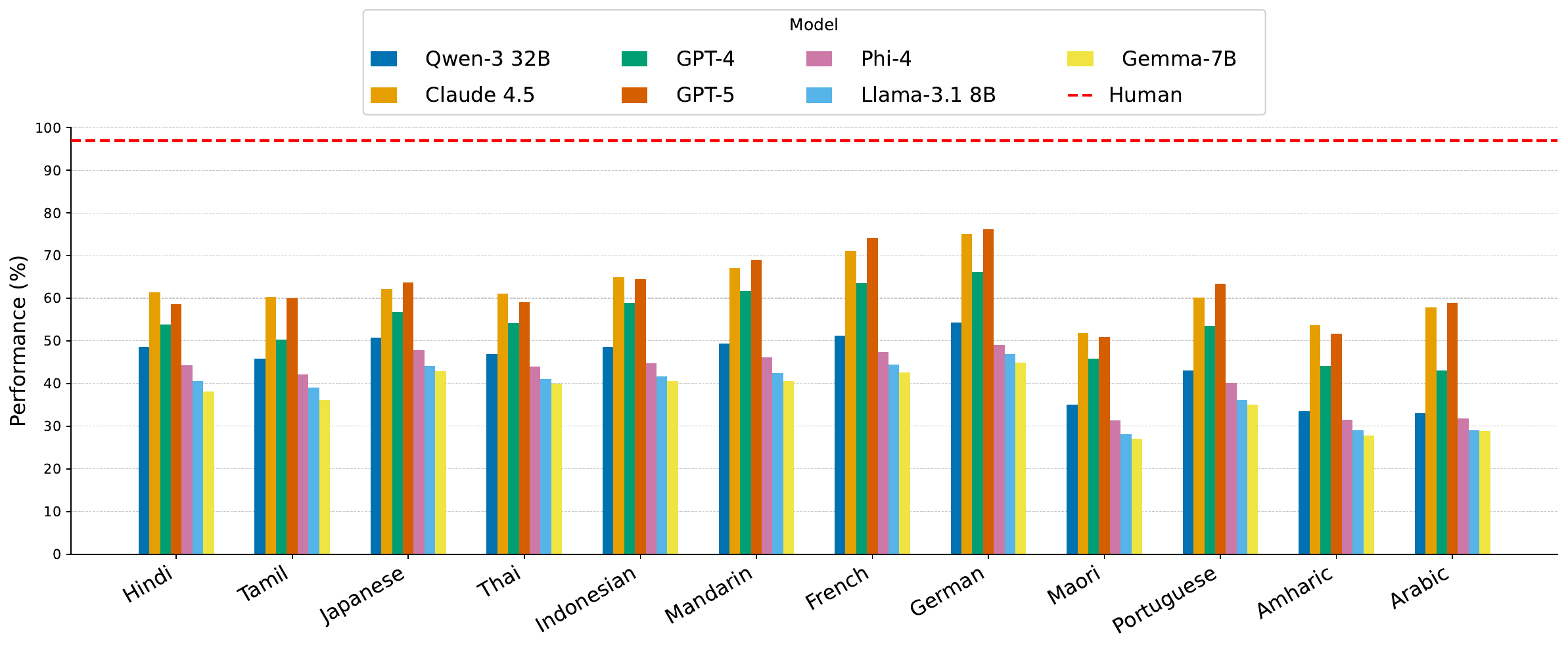}
\caption{Results of LLMs and SLMs on rule-based questions across languages (text modality).} 
\label{Reuleresult}
\end{figure*}

\begin{figure*}
\includegraphics[width=1\textwidth]{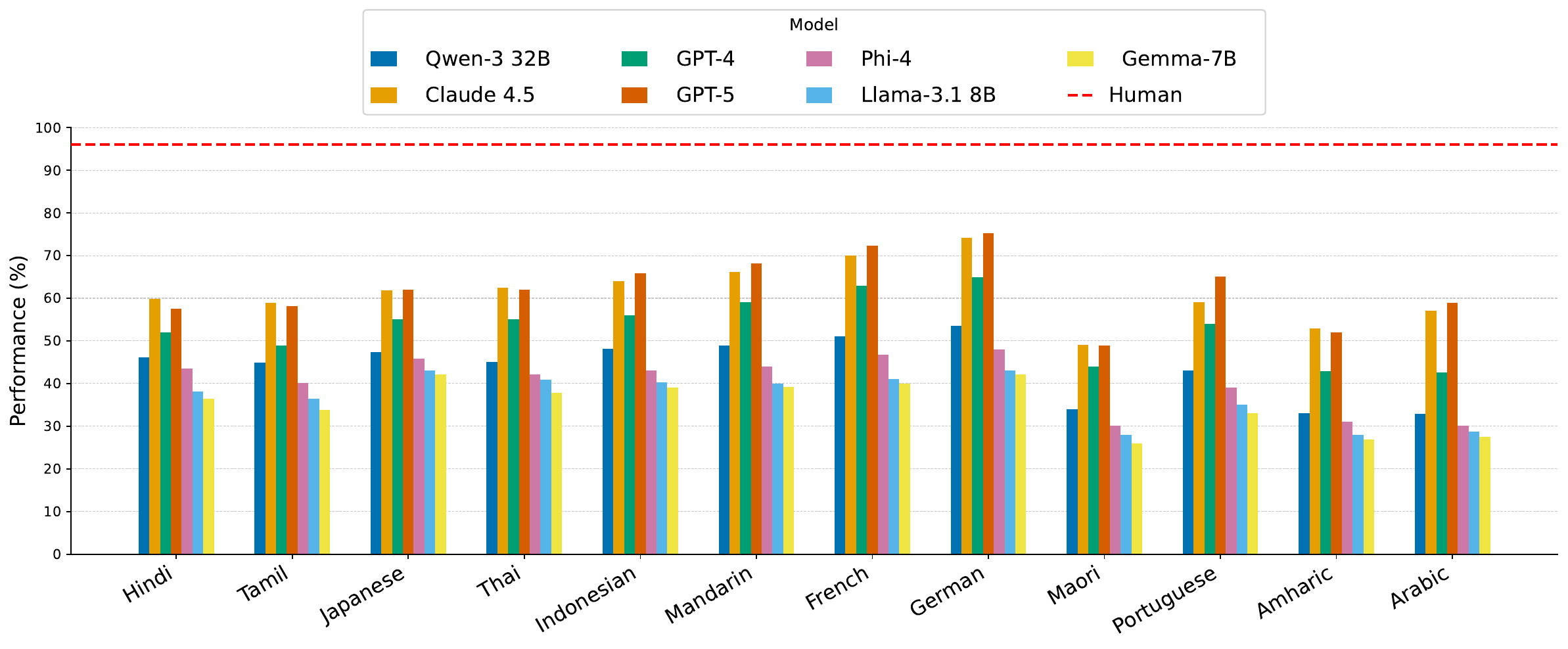}
\caption{Results of LLMs and SLMs on history-based questions across languages (text modality).} 
\label{HistoryResult}
\end{figure*}

\begin{figure*}
\includegraphics[width=1\textwidth]{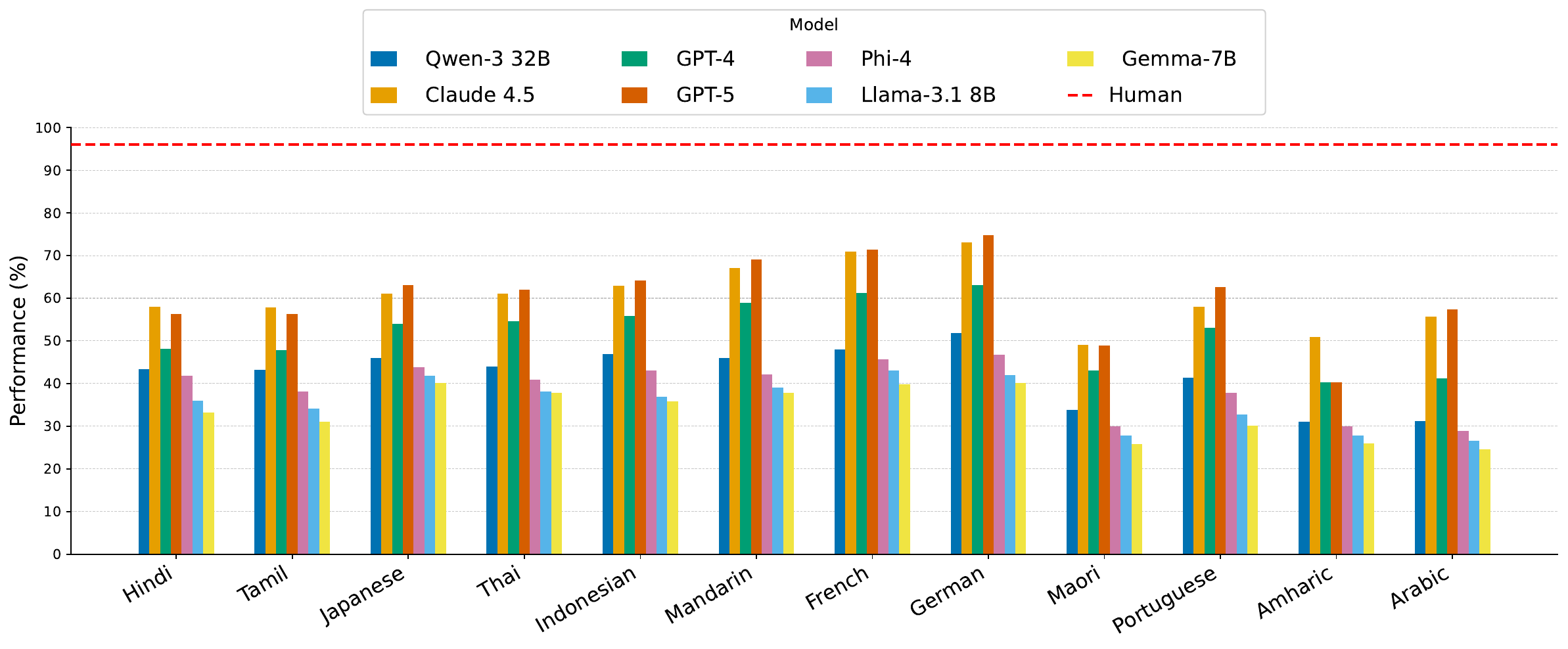}
\caption{Results of LLMs and SLMs on scenario-based questions across languages (text modality).} 
\label{scenarioSresult}
\end{figure*}

\textbf{Performance of LLMs and SLMs.} Across our multilingual evaluation, frontier LLMs consistently outperform compact models.   GPT-5 emerges as the top performer (61.73\%), closely followed by Claude Opus 4.5 (61.32\%).  Both display robust stability across categories: GPT-5 scores 62.51\% (Rule), 62.18\% (History), and 60.51\% (Scenario), while Claude-4.5 Opus yields 62.21\%, 61.29\%, and 60.46\%, respectively.  GPT-4 occupies a distinct secondary tier at 53.08\%. Among mid-range models, Qwen-3 32B (43.76\%)  outperforms Phi-4 (40.36\%), LLaMA-3.1-8B (36.99\%), and Gemma-7B (35.32\%).  Universally, models excel in European and East Asian languages (e.g., German, French, Mandarin) but degrade on low-resource languages (e.g., Māori, Amharic, Arabic). This highlights a persistent cross-lingual disparity, demonstrating that model scale correlates with better generalization on cultural QA tasks. Granular breakdowns are provided in Figures \ref{Reuleresult}, \ref{HistoryResult}, and \ref{scenarioSresult}.


\begin{figure*}
\includegraphics[width=1\textwidth]{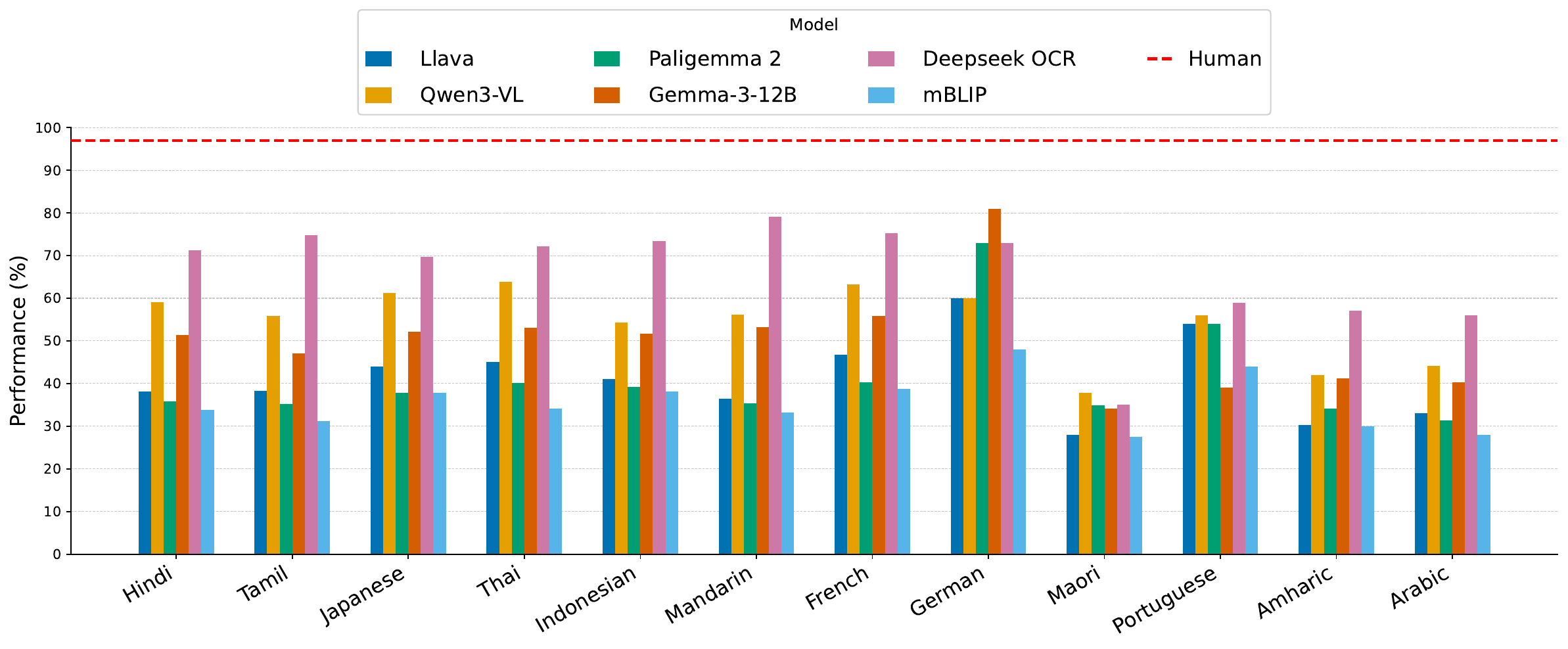}
\caption{Results of MLMs and SMLMs on rule-based questions across languages (image modality).} 
\label{ImageRuleResult}
\end{figure*}
\begin{figure*}
\includegraphics[width=1\textwidth]{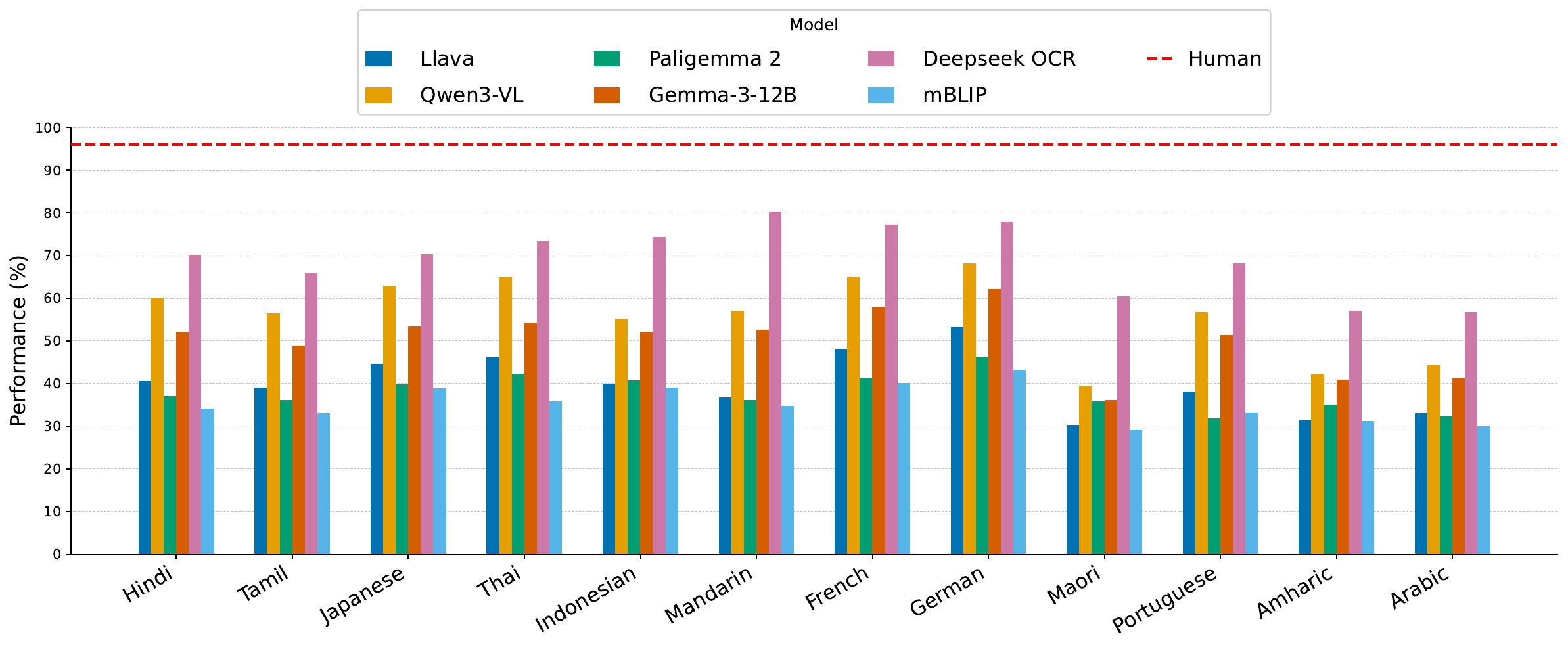}
\caption{Results of MLMs and SMLMs on history-based questions across languages (image modality).} 
\label{ImageHistoryResult}
\end{figure*}
\begin{figure*}
\includegraphics[width=1\textwidth]{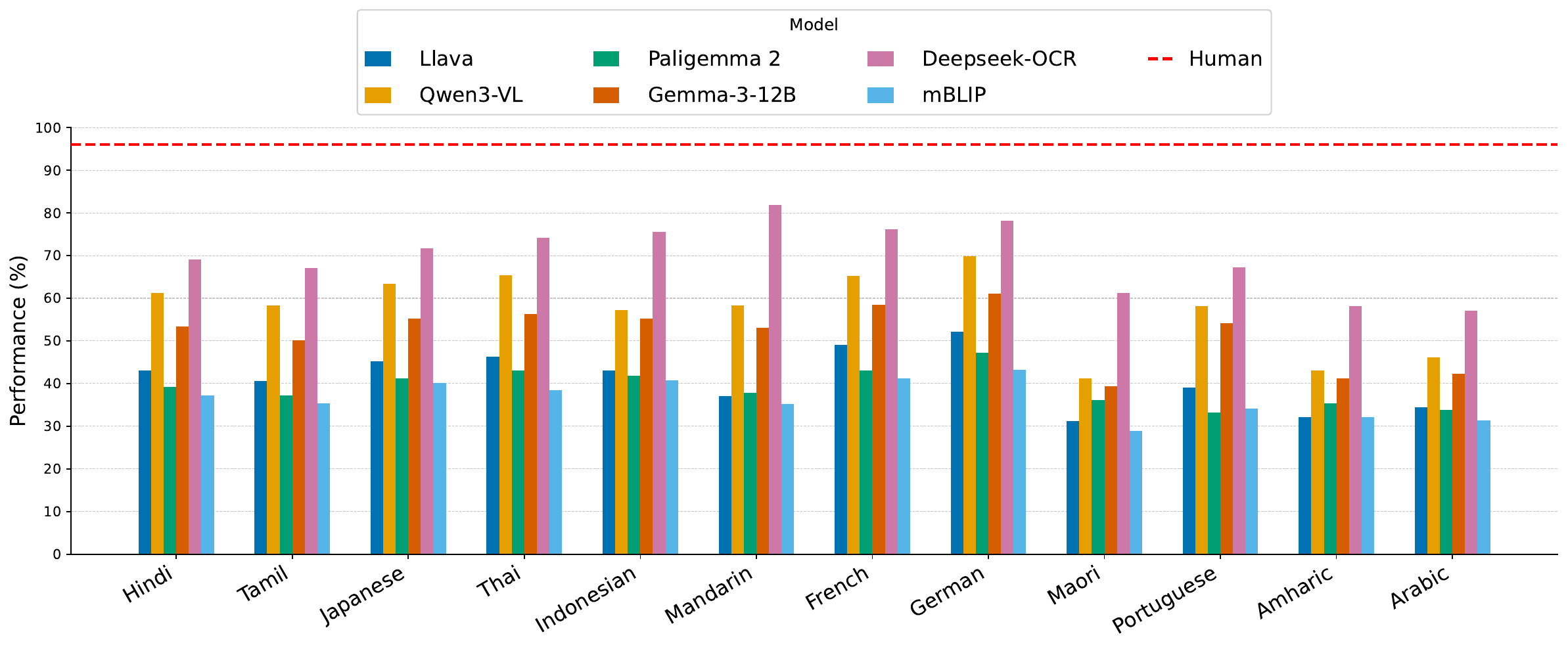}
\caption{Results of MLMs and SMLMs on scenario-based questions across languages (image modality).} 
\label{imageScenarioResult}
\end{figure*}

\textbf{Performance of MLMs and SMLMs.}  Across the multilingual multimodal evaluation, DeepSeek-OCR emerges as the top-performing system with $68.64\%$ overall average, consistently leading across Scenario-based ($69.80\%$), History ($69.32\%$), and Rule-based ($66.30\%$) categories.  Qwen3-VL forms a distinct secondary tier, achieving a solid overall average of $55.77\%$ (Scenario: $57.32\%$, History: $56.02\%$, Rules: $54.47\%$). Among smaller and mid-range architectures, Gemma-3-12B performs competitively at $50.73\%$, outperforming LLaVA ($40.95\%$), PaliGemma-2 ($40.26\%$), and mBLIP ($35.83\%$). Universally, models excel on scenario-based reasoning but encounter performance degradation on historical and rule-oriented cultural tasks, particularly under low-resource language settings. A shared cross-lingual bottleneck is observed: while architectures generalize effectively in high-resource languages (e.g., German, French, Mandarin, Japanese, and Thai), low-resource targets like Māori, Amharic, and Arabic present persistent challenges. These results confirm that advanced vision-language scales substantially improve multimodal cultural comprehension, whereas compact architectures continue to struggle with historically grounded and rule-driven reasoning. Figures \ref{ImageRuleResult}, \ref{ImageHistoryResult}, and \ref{imageScenarioResult} provide granular performance breakdowns.


\subsection{Human Evaluation}
Human raters evaluated the rule-based, history-based, and scenario-based categories to assess cultural reasoning. In the text modality, accuracies were 97\%, 96\%, and 96\%, while in the image modality, they were 97\%, 96\%, and 96\%, respectively. Despite strong human consistency, a partial misalignment with model predictions highlights that language models still fall short of replicating human-level cultural interpretation and contextual reasoning.

\section{Conclusion}
In this work, we construct \ourdata, a comprehensive benchmark designed to evaluate language models’ understanding of Asian, African, South American, Oceania, and European traditional dance. The dataset, consisting of 9,260 curated question-answer pairs from 12 countries, covers eight key aspects such as rules, history, location, attire, and other cultural significance.  Evaluations with leading models reveal notable gaps in answering traditional dance-specific questions, highlighting biases likely caused by training data limitations. \ourdata, built for quality and cultural sensitivity, advances inclusive AI research. Future expansions will add more languages and traditional dance to enhance its impact.

\section*{Limitations}
Despite being one of the most comprehensive evaluations of language models on traditional dance and cultural knowledge, this study has several limitations:
\begin{compactenum}[(1)]
    \item \textbf{Limited Geographic, Language, and Cultural Scope:} The \ourdata dataset currently covers 12 languages from 12 countries across 5 continents. While valuable, its scope remains limited.  Expanding to include more countries, especially those with underrepresented regions and low-resource languages, would improve cultural diversity and inclusivity, enabling fairer and broader assessments of language models.
    \item \textbf{Limited Representation of Traditional Dances:} Although the dataset covers 30 traditional dances—the largest multilingual and multimodal cultural dance dataset—it may still not fully capture the diversity of traditional dances across continents.  Future versions could include more dance forms and additional question types, such as True/False, adversarial, and scenario-based reasoning tasks.
\end{compactenum}

\section*{Ethics Statement}

\textbf{Data Collection and Bias Mitigation.} Data for \ourdata are collected from publicly accessible platforms (see Sec.~3.1), carefully selected to ensure authenticity. \ourdata thus represents a significant step toward a standardized, inclusive benchmark for traditional dances from Asia, Europe, South America, Oceania, and Africa.  All sources are verified by annotators through multiple group discussions, and irrelevant metadata was discarded.

\textbf{Human Annotation.} A diverse team of 36 annotators—experts in Asian, Oceanian, African, South American, and European dance, linguistics, and related fields—crafted, verified, and translated questions. The team included native and bilingual speakers from 12 countries, all with deep regional and dance knowledge. Annotators underwent comprehensive training on dataset objectives, question categories, and dance‑specific guidelines. Nearly all were native speakers with approximately 15 years of experience in dance, ensuring linguistic and cultural accuracy. The team, aged 28–50, provided a balanced generational perspective. Annotation was collaborative, with cross‑validation by a separate subteam to ensure consistency and address biases. Ethical considerations—avoiding stereotypes and promoting inclusivity—remained central, ensuring the dataset fairly reflects diverse cultural values.

\vspace{1cm}
\begin{credits}
\subsubsection{\ackname}
This work was supported by the Stable Support Program of Universities in Shenzhen (20231129211559001). We further acknowledge financial support from BMFTR and SMWK within the Center of Excellence for AI Research ``ScaDS.AI Dresden/Leipzig''.

\subsubsection{\discintname}
The authors have no competing interests to declare that are relevant to the content of this article.
\end{credits}
\bibliographystyle{splncs04}
\bibliography{Custom1}




\end{document}